\documentclass[sigconf, 9pt]{acmart}
\usepackage{hyperref}
\usepackage{graphicx}
\usepackage{caption}
\usepackage{adjustbox}
\usepackage{mathtools}
\usepackage{amsmath}
\usepackage{paralist}
\usepackage{multirow}
\usepackage[ruled, linesnumbered, noend]{algorithm2e} 
\usepackage[labelformat=simple]{subcaption}

\newcommand{\name}[0]{{\sc Babel}\xspace}
\newcommand{\eg}{{\it e.g.,}\xspace}
\newcommand{\ie}{{\it i.e.,}\xspace}

\copyrightyear{2025} 
\acmYear{2025} 
\setcopyright{acmlicensed}
\acmConference[SenSys 2025]{The 23rd ACM Conference on Embedded Networked Sensor Systems}{May 6-9, 2025}{Irvine, USA}
\acmPrice{}

\begin{document}
\title{\name: A Scalable Pre-trained Model for Multi-Modal Sensing via Expandable Modality Alignment}

\author{Shenghong Dai$^{\dagger}$}
\affiliation{%
  \institution{University of Wisconsin-Madison}
  \city{Madison}
  \state{WI}
  \country{USA}
}
\email{sdai37@wisc.edu}

\author{Shiqi Jiang}
\affiliation{%
  \institution{Microsoft Research}
  \city{Shanghai}
  \country{China}
}
\email{shijiang@microsoft.com}

\author{Yifan Yang}
\affiliation{%
  \institution{Microsoft Research}
  \city{Shanghai}
  \country{China}
}
\email{yifanyang@microsoft.com}

\author{Ting Cao}
\affiliation{%
  \institution{Microsoft Research}
  \city{Beijing}
  \country{China}
}
\email{ticao@microsoft.com}

\author{Mo Li}
\affiliation{%
  \institution{Hong Kong University of Science and Technology}
  \city{Hong Kong}
  \country{China}
}
\email{lim@cse.ust.hk}

\author{Suman Banerjee}
\affiliation{%
  \institution{University of Wisconsin-Madison}
  \city{Madison}
  \state{WI}
  \country{USA}
}
\email{suman@cs.wisc.edu}

\author{Lili Qiu}
\affiliation{%
  \institution{Microsoft Research}
  \city{Shanghai}
  \country{China}
}
\email{liliqiu@microsoft.com}



\renewcommand{\shortauthors}{Shenghong Dai, Shiqi Jiang, Yifan Yang, Ting Cao, Mo Li, Suman Banerjee, Lili Qiu}
\begin{abstract}
\label{sec:abstract}

This paper presents \name, the expandable modality alignment model, specially designed for multi-modal sensing. While there has been considerable work on multi-modality alignment, they all struggle to effectively incorporate multiple sensing modalities due to the data scarcity constraints. How to utilize multi-modal data with partial pairings in sensing remains an unresolved challenge.%

\name tackles this challenge by introducing the concept of \emph{expandable modality alignment}. The key idea involves transforming the N-modality alignment into a series of binary-modality alignments. Novel techniques are also proposed to further mitigate data scarcity issue and balance the contribution of the newly incorporated modality with the previously established modality alignment during the expandable alignment process. We provide the comprehensive implementation. In the pre-training phase, \name currently aligns 6 sensing modalities, namely Wi-Fi, mmWave, IMU, LiDAR, video, and depth. For the deployment phase, as a foundation model, any single or combination of aligned modalities could be selected from \name and applied to downstream tasks.%

Evaluation demonstrates \name's outstanding performance on eight human activity recognition datasets, compared to a broad range of baselines \eg the SOTA single-modal sensing networks, multi-modal sensing framework, and multi-modal large language models. \name not only improves the performance of individual modality sensing (12\% averaged accuracy improvement), but also effectively fuses multiple available modalities (up to 22\% accuracy increase). Case studies also highlight emerging application scenarios empowered by \name, including cross-modality retrieval (\ie sensing imaging), and bridging LLM for sensing comprehension. Our code is available at \href{https://aka.ms/project-babel}{aka.ms/project-babel}.

\end{abstract}

\thanks{$^\dagger$Research is done during internship at Microsoft Research}
\maketitle

\section{Introduction}
\label{sec:intro}

Sensing technology, with its distinctive capacity to perceive the physical world, has found widespread application across a multitude of domains, encompassing healthcare, mixed reality, smart driving, and beyond. Over the past several decades, a plethora of sensing modalities have been investigated, each offering a unique and complementary perspective of the physical world. This has led to the emergence of \emph{multi-modal sensing}, an approach that harnesses the simultaneous use of multiple sensing modalities.

Early methods for organizing multiple sensing modalities relied on handcrafted heuristics or features~\cite{zhouUseItFree2014}, which is proved challenging to scale across various modalities and tasks, due to the complexity of sensing signals and environments. Recent advancements in multi-modal learning have introduced promising solutions~\cite{zhaoSeethrough2018,DeldariXSSS22,JainTMKM22}. These methods automatically uncover correlations among diverse sensing modalities through supervised or self-supervised learning~\cite{ihianle2020deep,li2022deep,radu2018multimodal,chung2019sensor,salehi2022deep}. Among them, \emph{modality alignment} projects the representations of each sensing modality into a unified and shared space by leveraging paired modality data, demonstrating superior performance~\cite{ouyang2022cosmo}.

Although modality alignment can effectively organize sensing modalities, existing work~\cite{ouyang2022cosmo, radu2018multimodal,chung2019sensor,salehi2022deep} is often tailored for specific modalities, necessitating resampling and retraining for different downstream tasks and modality combinations, hindering seamless deployment of sensing applications. Therefore, this paper poses the question: \emph{Can we build a pre-trained multi-modal sensing alignment network as a foundation model}? This model would align common sensing modalities and allow for the integration of new modalities. In the deployment phase, any single or combination of aligned modalities from the model could be selected and applied to downstream tasks directly without retraining.

\begin{figure}[t]
    \centering
    \includegraphics[width=0.75\linewidth]{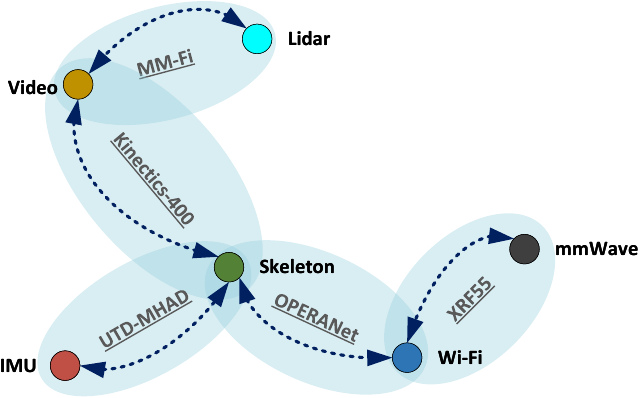}
    \caption{Five public sensing datasets, XRF55~\cite{10.1145/3643543}, OPERANet~\cite{bocus2021operanet}, MM-Fi~\cite{yang2023mmfi}, UTD-MHAD~\cite{chen2015utd} and Kinectics-400~\cite{kay2017kinetics} with binary-paired data cover six modalities.}
    \label{fig:expansive_modalities}
\end{figure}

While modality alignment in AI is a growing research area, its application in sensing presents significant hurdles. The fundamental challenge in supporting multi-modality in sensing is the \emph{data scarcity}, specifically, \textit{(i)} the scarcity of paired data, which is essential for aligning two modalities. For instance, the widely-used CLIP~\cite{radford2021learning} required 400 million image-text pairs for pre-training. In sensing, there lacks paired data from all modalities since some modality data require specialized hardware and expertise to collect. and \textit{(ii)} the scarcity of multi-paired modalities. Existing datasets only contain data from a subset of modalities~\cite{chen2015utd, yang2023mmfi, bocus2021operanet, kay2017kinetics}. For these reasons, existing research~\cite{han2023onellm, M4mengwei, ouyang2022cosmo, xu2023mesen, zhang2023metatransformer, girdhar2023imagebind} struggle to fully incorporate multiple sensing modalities. For instance, due to the limited language-paired data, OneLLM~\cite{han2023onellm} supports a limited number of sensing modalities, \ie IMU, with subpar performance (see Table~\ref{tab:performance_mllms}). Cosmos~\cite{ouyang2022cosmo} pioneered the alignment of multiple modalities, but due to the scarcity of multi-paired modalities, it aligns a limited number of modalities \eg IMU and depth.

In this paper, we present \name to address this challenge, establishing the first scalable pre-trained network aligning multiple sensing modalities. The design of \name is underpinned two observations: \textit{(i)} Despite the scarcity of paired data, there exist well-developed encoders or feature extractors for single modality sensing. By leveraging these encoders, the amount of paired data required for modality alignment could be significantly reduced. \textit{(ii)} Even though few datasets provide more than three paired modalities, numerous paired datasets exist that share common sensing modalities. These shared modalities can serve as a bridge for multi-modality alignment (see Fig.~\ref{fig:expansive_modalities}).

Drawing from these observations, the key idea of \name is the \emph{expandable multi-modal alignment}, which particularly transforms an N-modality alignment problem into a sequence of binary modality alignments. The \emph{expandability} facilitates the effective utilization of partially paired data in the sensing community. As illustrated in Fig.~\ref{fig:expansive_modalities}, we could achieve alignment of six modalities through five binary-modality alignments, using the corresponding datasets.

To realize this \emph{expandability}, we introduce three techniques: the pre-trained modality tower (\S\ref{sec:encoders}), the expandable network architecture (\S\ref{sec:expansive}), and the adaptive training strategy (\S\ref{sec:training_strategy}). Each modality utilizes a modality tower to extract features from raw data. We build these towers using existing singular-modal sensing feature extractors (\eg LIMU-BERT~\citep{xu2022limu} for IMU), and extent with our alignment modules for aligning with other modality towers; The expandable network architecture enables sequential training phases with only binary-paired samples. Within it we propose the prototype network, shared by all modalities, maintains the knowledge of aligned modalities when adding new ones. Lastly, our adaptive training strategy balances the contribution of newly added modalities to the unified representation, optimally assimilating new knowledge during model growth without disrupting established alignments.

We offer a comprehensive implementation of \name, including the network architecture, data preparation and processing, as well as the training details. In \name, we currently align six common sensing modalities: two for wireless sensing, namely Wi-Fi and mmWave, two for mobile sensing, specifically IMU and LiDAR, and two for general vision, namely RGB and depth. As an expandable framework, \name is allowed for aligning more modalities in the future without retraining aligned modalities.
In our work, five datasets are utilized to construct \name, including UTD-MHAD~\cite{chen2015utd}, Kinetics-400~\cite{kay2017kinetics}, OPERANet~\cite{bocus2021operanet}, XRF55~\cite{10.1145/3643543}, and MM-Fi~\cite{yang2023mmfi}.

The current pre-trained \name is evaluated on a typical sensing application, Human Activity Recognition (HAR), across eight datasets, which include both in-domain and out-of-domain datasets~\cite{chen2015utd, yang2023mmfi, bocus2021operanet, 10.1145/3643543, reyes2016transition, 9516988, an2022mrimultimodal3dhuman, li2010action}. To demonstrate \name's capability, we compared it to an array of baselines, including the state-of-the-art (SOTA) singular-modal sensing networks~\cite{xu2021limu, yang2023sensefi, li2010action, an2022mri}, multi-modal sensing framework~\cite{ouyang2022cosmo},  and the emerging Multi-modal Large Language Models (MLLMs)~\cite{han2023onellm, M4mengwei, girdhar2023imagebind, zhang2023metatransformer}.

We use the one-shot learning\footnote{For each HAR class, we just use only one sample to fine-tune the downstream task classification header.} to evaluate \name as a foundation model. The evaluation shows that, \textit{(i)} owing to the pre-trained alignment across more sensing modalities and extensive datasets, \name improve the accuracy of \emph{single modality sensing} by up to 20\%, and 12\% on average on all six aligned modalities across various datasets. \textit{(ii)} Thanks to the aligned unified embedding space, \name increases \emph{multi-modal sensing fusion} accuracy by up to 22\% compared to current multi-modal frameworks~\cite{ouyang2022cosmo}. \textit{(iii)} When comparing with emerging MLLMs,  which are limited in the range of sensing modalities they support, \name surpasses them by the accuracy of 25.2\% across HAR datasets.%

Besides HAR, we also present two application studies to highlight \name's potential as a foundation model. The first is \textit{sensing imaging} to illustrate cross-modality retrieval. With \name, the original image-to-image diffusion model can be supplemented with non-visual data as input to generate images~\citep{ramesh2022hierarchical}. The other case aims to bridge the gap between LLM and sensing. By injecting the IMU sensing signal through \name into the Video-LLaMA~\citep{zhang2023videollama}, the LLM can understand the sensing signals without any retraining.

To summarize, the contributions of the paper include: 
\begin{itemize}
    \item \name, to the best of our knowledge, is the first expandable foundation model for the multi-modal sensing, currently aligning six sensing modalities.

    \item Within \name, we introduce key techniques for learning with scarce paired sensing data and modalities, including the pre-trained modality tower, expandable network architecture, and adaptive training strategy.

    \item We demonstrate \name's superior performance compared to a range of baselines. Additionally, we highlight \name's potential in the field of cross-modality retrieval, and its ability to bridge LLMs for enhanced comprehension of the physical world.
\end{itemize}

\section{Related Work}

In this section, we introduce the work pertaining to the design of \name, and highlight our distinctive contribution divergent from the existing research. Specifically, we discuss key advancements in modality alignment, multi-modal sensing, and multi-modal LLMs.

\textbf{Modality alignment}, as an emerging research topic, involving various methods~\citep{DBLP:journals/corr/abs-1807-03748,DBLP:journals/corr/abs-1906-05849}. Of these, contrastive learning (CL) is notable. CL, a self-supervised learning method, differentiates similar and dissimilar samples by comparing positive (similar) and negative (dissimilar) pairs. The goal is to generate representations where similar samples are close, and dissimilar ones are far apart in the feature space. Contrastive Language-Image Pretraining (CLIP)~\cite{radford2021learning} exemplifies the effective use of CL in aligning text and image modalities. CLIP, trained on a large Internet image-caption pairs, learns to associate semantically related texts and images. FOCAL~\citep{liu2024focal} introduces an innovative CL framework with a temporal structural constraint designed for sensing data, addressing challenges in capturing both shared and modality-exclusive features in multimodal time-series data.

Nonetheless, it is still challenging to apply CL to align multiple sensing modalities, due to the data scarcity issue. For instance, CLIP's training necessitates approximately $400,000,000$ image-text pairs, a scale of data that public datasets with paired sensing samples fail to match. In fact, public multi-modal sensing datasets~\citep{chen2015utd,an2022mri,chen2023mmbody}, as an example, contain a mere $600-42,000$ sample pairs, a stark contrast to the required volume. Additionally, there exists numerous sensing modalities. The alignment of \emph{N} sensing modalities generally necessitates a substantial amount of \emph{N-tuple} data. Regrettably, There are no public datasets that cater to the alignment of a greater number of sensing modalities, such as six or more. \name addresses this fundamental challenge via the proposed \emph{expandable modality alignment} technique.

\textbf{Multi-modal sensing} offers unique abilities to perceive the physical world, incorporates a plethora of methods. For instance, Cosmo~\cite{ouyang2022cosmo} pioneered the application of contrastive fusion learning in multi-modal sensing, incorporating RGB, depth and IMU modalities. MESEN~\cite{xu2023mesen} employs multi-modal contrastive learning to improve the performance of singular-modal sensing. FM-Fi~\citep{weng2024largemodelsmalldata} leverages CLIP through cross-modal contrastive knowledge distillation to improve Radio-Frequency-based human activity recognition with limited labeled data. Nevertheless, these studies are typically crafted for chosen modalities, necessitating retraining for additional ones. In stark contrast to Cosmo, MESEN and FM-Fi, \name operates as a pre-trained foundational model for multi-modal sensing, facilitating the utilization of any single or multiple aligned modalities for downstream sensing tasks without retraining. Moreover, due to the pre-trained alignment across a broad spectrum of modalities, \name attains exceptional performance in both single-modal sensing and multi-modal fusion (\S\ref{sec:evaluation}).

The concept of multi-modal sensing is extensively employed across a wide array of applications. For instance, \citep{liu2021rfid} integrates RFID and RGB for recognizing human-object interactions. \citep{wang2022multi} leverages LiDARs, cameras, and IMU and GNSS devices worn by animals to recognize animal behavior. To locate target individuals, \citep{10.1145/3458864.3466904} utilizes Wi-Fi Fine Timing Measurements and IMU data to associate individuals in a video with a matched query ID. GaitVibe+ \citep{Dong_2022} enhances structural vibration-based footstep localization using temporary cameras and vibration sensors for in-home gait analysis. \citep{girod2001robust} presents an acoustic and camera sensing system that ameliorates range estimation for applications in robotics and others. These applications could benefit through \name.

\textbf{Multi-modal LLMs} are rapidly evolving to accommodate an increasing number of modalities. Supporting an expanded range of modalities typically necessitates a multi-modal encoder, which projects multi-modal signals into the language embedding space. To construct such an encoder, Meta-Transformer~\citep{zhang2023metatransformer} demonstrates the potential of employing a shared transformer encoder across 12 modalities. ImageBind~\citep{girdhar2023imagebind} aligns six modalities utilizing solely image-paired data. Similarly, LanguageBind~\citep{zhu2024languagebindextendingvideolanguagepretraining,zhou2023tentconnectlanguagemodels} employs the language as the central binding modality to align four modalities. OneLLM~\citep{han2023onellm} aligns eight modalities to language using a singular, unified encoder. CoDi~\citep{tang2023anytoanygenerationcomposablediffusion} facilitates alignment across language, image, video, and audio modalities.

These studies, however, offer exceedingly limited support for sensing modalities. Indeed, the sole supported sensing modality is the IMU, yet its performance is notably subpar (see the detailed evaluation results in \S\ref{sec:evaluation}).  This is primarily due to the high requirement for data with specific modality pairings, such as image-sensing pairs. For instance, ImageBind~\citep{girdhar2023imagebind} is only trained on Ego4D~\citep{grauman2022ego4d} for IMU. Their cross-domain capabilities are restricted and they cannot be trained on other sensing modalities without first addressing the issue of data scarcity. In response, we propose crucial techniques for aligning sensing modalities through an expandable architecture, reducing dependence on exhaustive modality pairings. Details are presented in the following.

\section{\name Overview}
\label{sec:overview}

\begin{figure}[t]
    \centering
    \includegraphics[width=\linewidth]{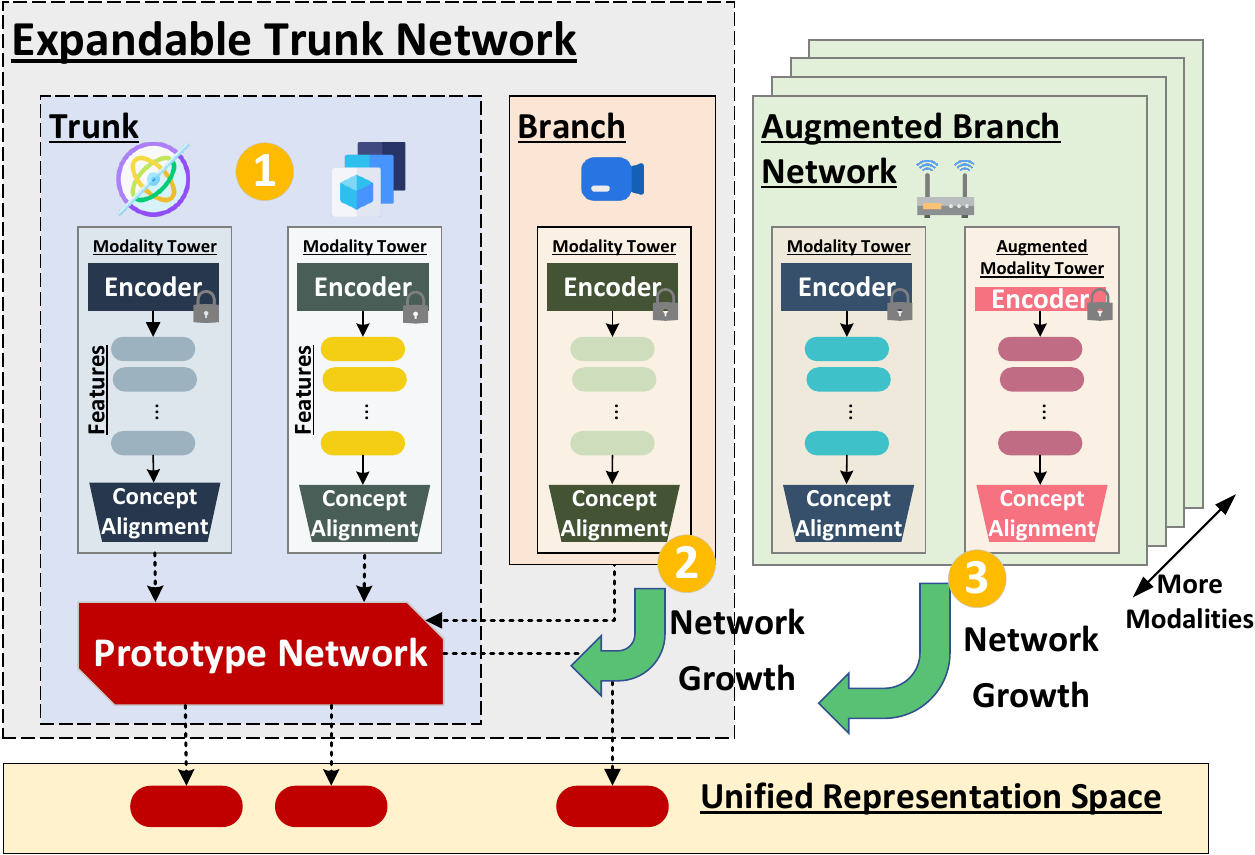}
    \caption{Overview of \name.}
    \label{fig:overview}
\end{figure}

\name, to the best of our knowledge, is the first scalable multi-modal pre-trained network, specifically designed for sensing applications, suitable for a multitude of downstream tasks. \name consists of the model architecture designs, training strategies as well as the data preparation and processing techniques.
In \name, we present two designs to build the network with constraint data, namely \emph{pre-trained modality tower} and \emph{expandable model architecture } to cope with the scarcity challenge of paired sensing data and multi-paired sensing modalities.

In the design of the \textbf{pre-trained modality tower}, our aim is to harness the power of existing feature extractor within singular modality sensing to construct the modality alignment network, thereby significantly decreasing the necessity for extensive paired training samples.

The crux of this design lies in the efficient alignment of representations across pre-trained encoders. Thereby, we introduce the \emph{modality tower}, consisting of the pre-trained encoder, and the concept alignment module.
The encoder could be based on signal processing and neural networks from existing deep learning models. 
The concept alignment module then aligns embeddings (features) from encoders. During training, pre-trained encoders are frozen, and the concept alignment module is updated.

In the design of the \textbf{expandable model architecture}, we try to convert the contrastive training process with \emph{N-tuple} samples into a sequence of training phases involving only paired samples, thereby reducing the need for tupled samples, rendering the alignment of multiple modalities truly feasible.

As illustrated in Fig.~\ref{fig:overview}, we initially align two modalities to form a \emph{trunk} network. We then introduce a new \emph{branch} modality, identifying the \emph{junction} modality within the trunk that pairs with the branch according to available training samples. Through CL, the branch is merged with the trunk to form the updated trunk, by aligning the branch and junction modality. We refer this process as \emph{growth}. The crux of this design lies in effectively maintaining knowledge of aligned modalities while assimilating new insights from the newly merged modality. Thereby, we introduce \emph{prototype network}, which is shared by all modalities and is carefully updated during training with our \textbf{adaptive training} strategy.

Adaptive training strategy is explicitly engineered for the sensing modality alignment. Particularly, during each training phase, we aim to create an embedding space where similar sample pairs converge by adjusting each modality's representation. The adjustment weights are vital as modalities contribute differently to the final space. More weight should be given to modalities with more clear signals, while those with more noise or fewer insights should contribute less, to preserve aligned modalities' knowledge. This balance varies depending on the modality combinations, datasets, and tasks. Hence, we propose an adaptive strategy for automatically determining weights.

Next we would introduce these designs in detail.

\section{Pre-trained Modality Tower}
\label{sec:encoders}

\subsection{Assembling Modality Towers}

In the alignment of each modality, our initial step involves constructing a \emph{modality tower}. Subsequent to this, we execute the contrastive learning on these modality towers. The modality tower incorporates two fundamental components: a pre-trained encoder, and a concept alignment module.

Comparing with conventional modality alignment methods \ie CLIP, \name's key design lies in the utilization of a pre-trained encoder within a singular modality, proving particularly effective for sensing modalities. Sensing modalities (e.g., IMU, LiDAR, Wi-Fi) have matured over decades of research, leading to specialized feature extractors. These encoders incorporate domain knowledge in their architecture, training scheme, or signal-processing pipeline. By reusing these encoders rather than designing new ones from scratch, \name benefits from the high-quality, domain-specific representations already learned in prior work.

\name's effectiveness 
can be attributed to two key factors. Firstly, the process of assembling the modality tower adheres to the proven method of parameter-efficient fine-tuning (PEFT)~\citep{liu2022fewshot}, a technique notably successful in addressing the vision-language modality alignment problem, as evidenced by models like LiT~\citep{zhai2022lit} and APE~\citep{rosenfeld2022ape}. The concept alignment module could be regarded as an \emph{adapter} in the context of PEFT practices. Secondly, the successful application of PEFT necessitates that the encoders can capture generic features. For modalities such as vision and language, it typically demands pre-training on a substantial corpus of data, ensuring that the pre-trained model does not exhibit significant domain shift and adequately covers representative features for a majority of downstream tasks.

Pertaining to sensing modalities, the input signals are typically modulated, bearing distinct physical interpretations, thereby making them distinctly defined and explicable in terms of physics. 
As sensing techniques advance, these representative features are further amplified. 
As a result, we note that the representative features of sensing modalities for a multitude of downstream tasks often remain consistent. 
This consistency facilitates our opportunity to leverage singular modality encoders in constructing the modality tower, following the practice of PEFT.

The particular encoder for each modality is chosen based on the following criteria. For modalities dedicated to sensing tasks, such as mmWave, we tend to choose signal processing-based encoders, due to their capability to extract universally applicable features with well-defined physical meanings. For more ubiquitous sensing modalities, like Wi-Fi, which are often noisy, we lean towards deep learning (DL)-based encoders, owing to their proficiency in de-noising. We avoid choosing models whose pre-training corpus might exhibit excessive domain shift compared to typical multi-modal sensing tasks. For example, a LiDAR encoder trained solely on autonomous-driving road scenes might not generalize well to indoor human-tracking data. Conversely, encoders trained on more general sets (e.g., broad 3D shapes for LiDAR, a diverse set of action videos) yield a feature space that better suits the variety of tasks \name must handle.
For modalities with large variation or especially noisy signals (e.g., Wi-Fi), relying on a single pre-trained encoder can be limiting. In such cases, we introduce modality tower augmentation, where multiple encoders are employed for the same modality, as elaborated in \S\ref{sec:aug_modality}.
Eventually we evaluate and compare the selected candidates by fine-tuning and testing them on a variety of singular modality datasets. The encoder demonstrating superior generality is chosen. To align modality embeddings, we employ a MLP layer as the concept alignment module. MLP-based projection layers have been widely used in contrastive learning frameworks~\citep{chen2020simple}.

\begin{figure}[!t]
    \centering
    \includegraphics[width=0.75\linewidth]{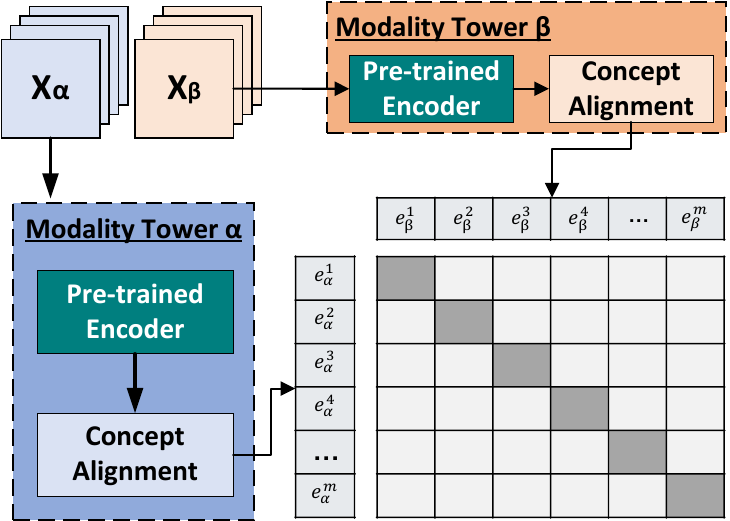}
    \caption{The alignment of two modality towers with pre-trained encoders and concept alignment modules, where \( e^m_{\alpha} \) denotes the \( m \)-th embedding of modality \( \alpha \).}
    \label{fig:align_towers}
\end{figure}

\subsection{Aligning Modality Towers}
\label{subsec:align}

Upon assembling the modality tower for a given modality, we strive to align them through the contrastive learning. Next, we would illustrate our modality alignment process using the alignment of two modalities as an example. The alignment of multiple modalities would be discussed in \S\ref{sec:expansive}.

As illustrated in Fig.~\ref{fig:align_towers}, given the dataset $E_{\alpha\beta}$ comprising paired samples of modality $\alpha$ and modality $\beta$, our first step is to structure the positive pairs $P$ and negative pairs $Z$ essential for the contrastive learning process. Specifically, the dataset $E_{\alpha\beta}$ includes sample pairs $(\chi_\alpha, \chi_\beta)$ that are initially synchronized. For instance, in the UTD-MHAD dataset~\citep{chen2015utd}, Each sample pair signifies a sequence of IMU readings and a concurrent video recording series of the same human activity, captured within a span of 5 seconds. From the dataset $E_{\alpha\beta}$, we randomly select a batch $M$ comprising $m$ sample pairs. Within this batch, for a given sample of modality $\alpha$, denoted as $\chi_\alpha^i$ where $i \in N$, we construct its corresponding positive pair $P_\alpha^i$ and negative pairs $Z_\alpha^i$ in the following manner,
\begin{align}    
    P_\alpha^i &= (\chi_\alpha^i, \chi_\beta^i), 1 \leq i \leq m, \\
    Z_\alpha^i &= \{(\chi_\alpha^i, \chi_\beta^j)\}, 1 \leq i,j \leq m, i \neq j,
\end{align}

Likewise, we can construct the positive pair $P_\beta^i$ and negative pairs $Z_\beta^i$ for the $i$th sample of modality $\beta$ within the batch $M$. Ultimately, for the batch $M$ consisting of $m$ pairs, we could derive $m$ positive pairs and $m^2-m$ negative pairs, which will be utilized in the sequential contrastive learning.

Throughout the training phase, the assembled positive pairs $P$ and negative pairs $Z$ are processed through the modality tower. 
The contrastive loss $L$ is computed on a per-batch basis for each batch $M$,

\begin{equation}\label{equ:loss}
    L_{\alpha\beta}^M = \frac{L_{\alpha\leftarrow\beta}^M + L_{\beta\leftarrow\alpha}^M}{2}, 
\end{equation}

where $L_{\alpha\leftarrow\beta}^M$ and $L_{\beta\leftarrow\alpha}^M$ denote the computed contrastive loss transitioning from modality $\beta$ to modality $\alpha$ and vice versa within the batch $M$, as defined subsequently,

\begin{equation}\label{equ:sim}
    L_{\alpha \leftarrow \beta}^M = -\sum_{i=1}^{m} \log \left( \frac{\exp (\text{sim}(P_\alpha^i) / \tau)}{\sum \exp (\text{sim}(N_\alpha^i) / \tau)} \right),
\end{equation}

where $\tau$ is a temperature parameter employed to scale the logits. In our implementation, we set $\tau$ to 0.07. The function $sim$ represents the cosine similarity function utilized to examine the output embeddings from $\Gamma_\alpha$  and $\Gamma_\beta$. Similarly, we can compute $L_{\beta\rightarrow\alpha}^M$. Eventually we use $L_{\alpha\beta}^M$ to update the concept alignment modules of modality towers of $\alpha$ and $\beta$.

As a pre-trained network, when \name is incorporated into downstream tasks,
we would introduce an additional task-specific network. For instance, a classifier head is introduced for activity classification tasks. Owing to the modality alignment, the aligned embedding from each modality can be straightforwardly concatenated for downstream tasks. As will be demonstrated in the evaluation, the output embeddings, enhanced by modality alignment, are significantly superior. Consequently, we can attain SOTA results even with a very simple classifier, such as a 2-layer MLP, when only applying one-shot learning.

\subsection{Augmenting Modality Towers}\label{sec:aug_modality}

We also propose to augment the modality towers by employing multiple encoders for the particular modality.
The concept of modality tower augmentation is inspired by model ensembling~\citep{breiman1996bagging,caruana2004ensemble}, where multiple weak learners combine to create a stronger one, improving accuracy and performance. This method has proven to effectively decrease variance and bias in each weak learner.

In \name, we would construct an augmented modality tower when incorporating additional encoder. We align the augmented modality towers in accordance with the process delineated in~\ref{subsec:align}.
Specifically, We construct two modality towers, $\Gamma_\alpha^\epsilon$ and $\Gamma_\alpha^\eta$, using pre-trained encoders $\epsilon$ and $\eta$ respectively. We align these towers using positive pairs $P_\alpha^i=(\chi_\alpha^i, \chi_\alpha^i)$ and negative pairs $Z_\alpha^i=\{(\chi_\alpha^i, \chi_\alpha^j)\}$ where $i\neq{j}$. The alignment is achieved through loss functions from Equations~\ref{equ:loss} and~\ref{equ:sim}. The similarity $sim$ is computed using output embeddings from both towers.

\begin{figure}[!t]
    \centering
    \includegraphics[width=0.75\linewidth]{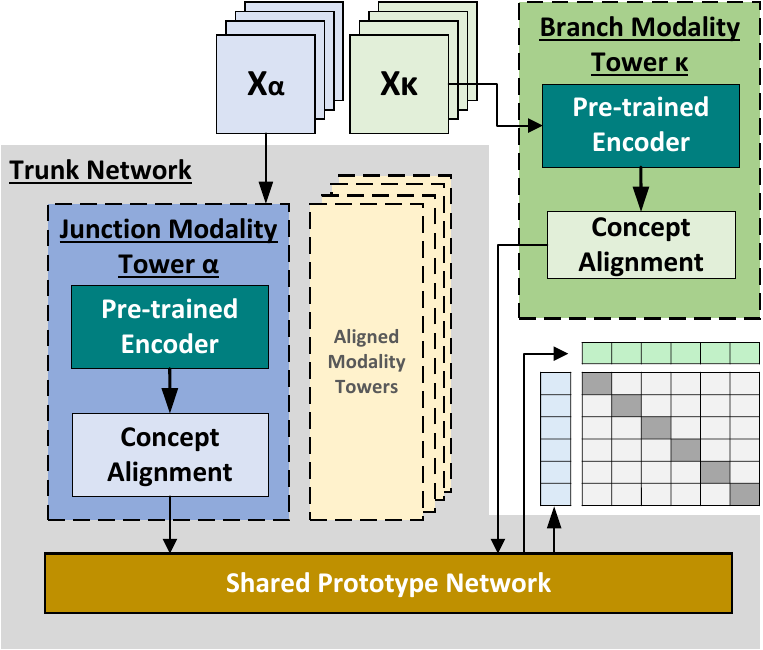}
    \caption{The alignment of multiple modalities with the prototype network in the expandable network architecture.}
    \label{fig:network_growth}
\end{figure}

\section{Expandable Model Architecture}
\label{sec:expansive}

\begin{figure*}[t!]
    \centering
    \begin{subfigure}{0.23\textwidth}
        \includegraphics[width=\linewidth]{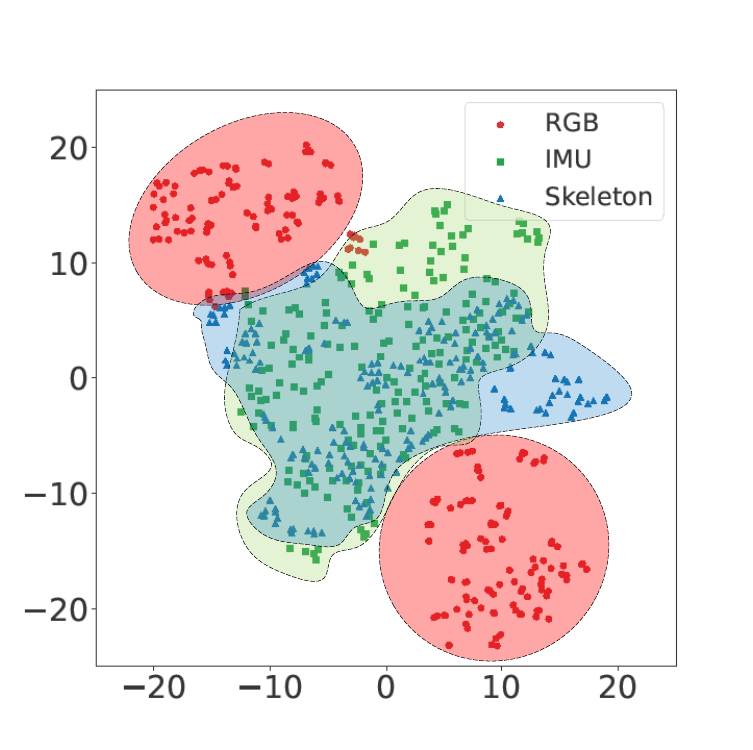}
        \caption{Without alignment}
        \label{subfig:before_align}
    \end{subfigure}
    \hfill
    \begin{subfigure}{0.23\textwidth}
        \includegraphics[width=\linewidth]{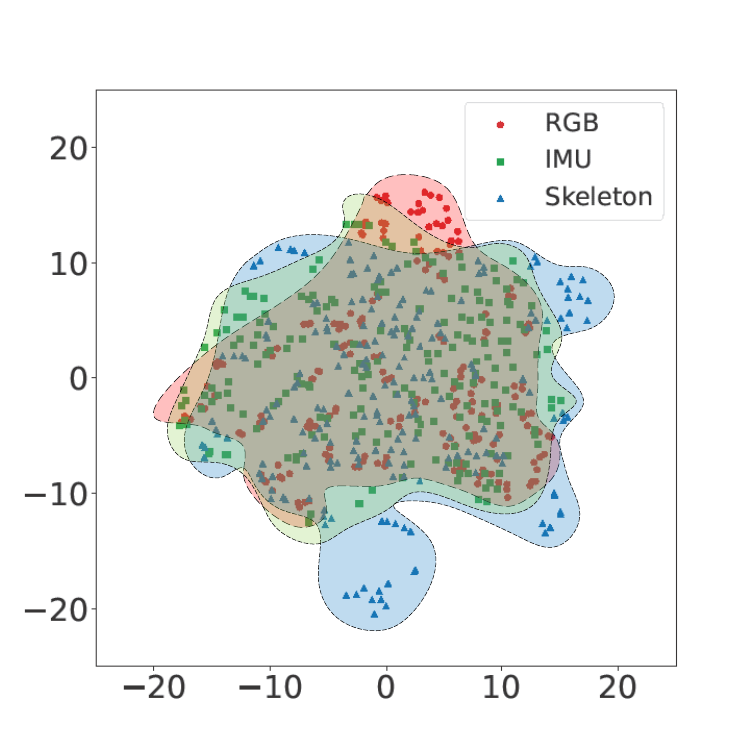}
        \caption{Triplet alignment}
        \label{subfig:joint_align}
    \end{subfigure}
    \hfill
    \begin{subfigure}{0.23\textwidth}
        \includegraphics[width=\linewidth]{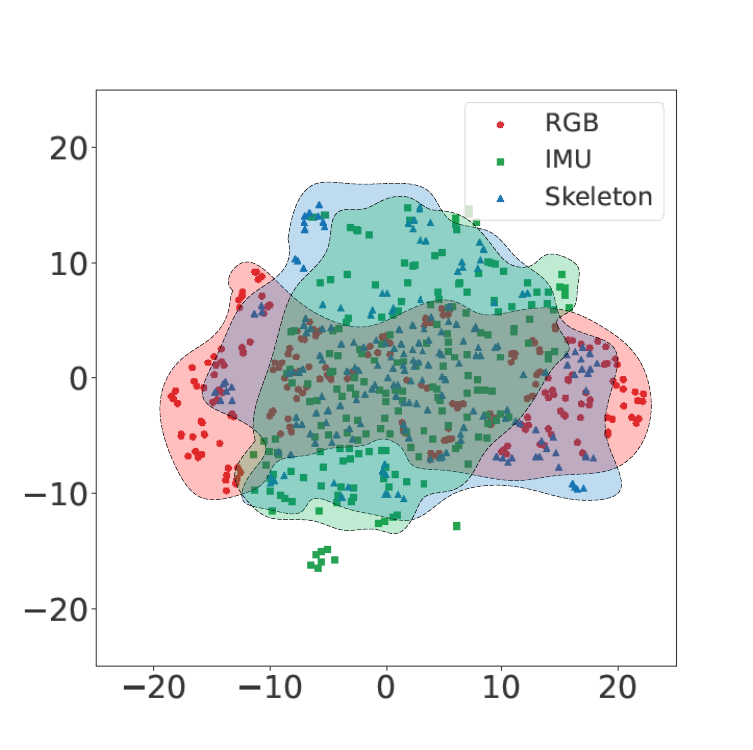}
        \caption{\name \footnotesize{(IMU-Skeleton-Video)}}
        \label{subfig:our_align_1}
    \end{subfigure}
    \hfill
    \begin{subfigure}{0.23\textwidth}
        \includegraphics[width=\linewidth]{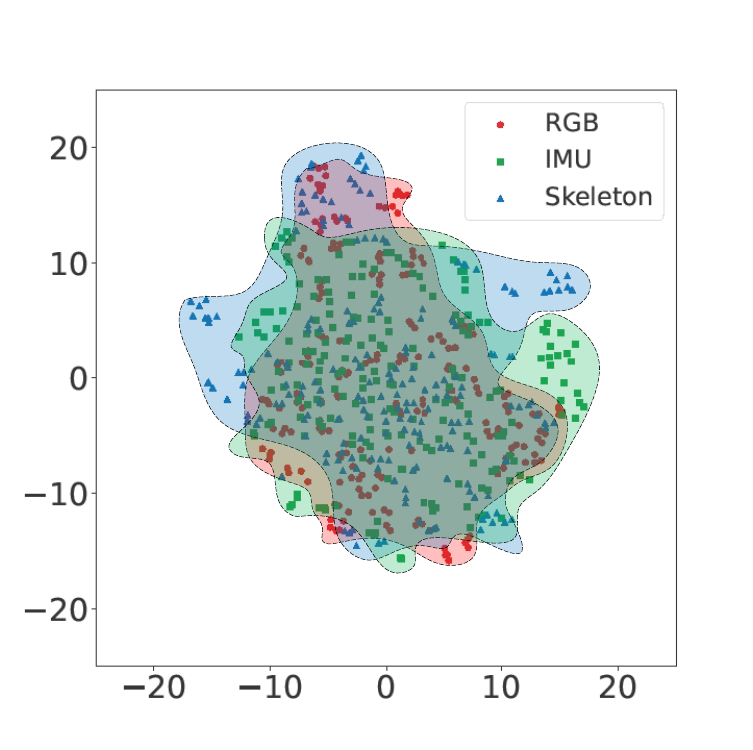}
        \caption{\name \footnotesize(Skeleton-Video-IMU)}
        \label{subfig:our_align_2}
    \end{subfigure}
    \caption{t-SNE representations of three modalities obtained by different modality alignment approaches.}
    \label{fig:representation}
\end{figure*}

\subsection{{Prototype Network}}
Aligning multiple sensing modalities (such as six or more) with partially paired modalities is challenging. In response to this, one of key designs in \name is the expandable model architecture, which transforms the training process for $N$ modality alignment into a series of two modality alignment phases, exploiting existing datasets with paired modalities.

To elaborate, consider the alignment of three modalities: $\alpha, \beta, \kappa$, with the available datasets $E_{\alpha\beta}$ and $E_{\alpha\kappa}$. We initially employ $E_{\alpha\beta}$ to align the modalities $\alpha$ and $\beta$, as discussed in \S\ref{subsec:align}, yielding the network $H_{\alpha\beta}$, which we term the \emph{trunk} network. Subsequently, we aim to integrate an additional modality $\kappa$ into the trunk $H_{\alpha\beta}$. 

Given that dataset $E_{\alpha\kappa}$ provides corresponding pairs between the modalities $\alpha$ and $\kappa$, we designate $\alpha$ as the \emph{junction} modality. From the trunk $H_{\alpha\beta}$, we select the trained modality tower $\Gamma_{\alpha}$. We then construct a new modality tower $\Gamma_{\kappa}$, referred to as the \emph{branch}. This branch is integrated into the trunk network by aligning the junction modality tower $\Gamma_{\alpha}$ with the branch modality tower $\Gamma_{\kappa}$, utilizing samples from the dataset $E_{\alpha\kappa}$. We refer to this procedure as network \emph{growth}. Fig.~\ref{fig:network_growth} illustrates the network growth in our expandable network architecture.

The challenge of facilitating network growth lies in maintaining the knowledge of previously aligned modalities while concurrently assimilating new insights from the additional modality. Therefore, 
during this growth phase, it is not suitable to directly align $\Gamma_{\alpha}$ and $\Gamma_{\kappa}$ as outlined in \S\ref{subsec:align}, since any updates to the junction modality $\Gamma_{\alpha}$ may significantly disrupt the already aligned modalities, such as modality $\beta$.

To this end, we introduce the prototype network. As shown in Fig.~\ref{fig:network_growth}, it is specifically incorporated into the trunk network, succeeding the concept alignment module of each modality tower. The prototype network is shared across all modality towers within the trunk network. It serves as a coordinating entity for all the learned knowledge across aligned modalities. Therefore by adjusting the updates on the prototype network, we could strike a balance between acquiring new knowledge from the branch modality and avoiding catastrophic forgetting of the trunk network.

Revisiting our previous example, during the initial alignment of modalities $\alpha$ and $\beta$, we concurrently update the prototype network $\Upsilon$ while training the concept alignment module of the modality towers $\Gamma_{\alpha}$ and $\Gamma_{\beta}$. Subsequently, during the network growth phase involving the branch modality $\kappa$ and the junction modality $\alpha$, the contrastive learning process would updates the branch and junction modality tower $\Gamma_{\kappa}$ along with the prototype network $\Upsilon$.

In our implementation, the structure of the prototype network is kept relatively straightforward, resembling a 2-4 layer MLP. Despite its simplicity, this design enables several advantages for the alignment of multiple modalities. First, during each network growth phase, it allows us to utilize different datasets, even for disparate tasks. Second, this design facilitates the repeated enhancement of aligned modalities using varied datasets. By assimilating insights from these different datasets, it becomes feasible to construct a pre-trained network with substantial generality. 

Together with the prototype network, we also devise the adaptive training strategy to regulate the extent to which the trunk network acquires new knowledge, which would be discussed in \S\ref{sec:training_strategy}.

\subsection{{Growth Orders}}
\name transforms the $N$-tuple modality alignment into a sequence of two-modality alignment phases, thereby raising a potential question regarding the differences between the conventional completed alignment and our expandable alignment with varying modality growth orders. The  insight of our prototype network is that it maintains a shared set of parameters across all modalities, inherently encoding common features learned from previous alignment phases. When aligning a new branch modality with the junction, the prototype network is updated in a way that partially shifts the shared embedding space, but does not overwrite it completely. This mitigates catastrophic forgetting~\cite{french1999catastrophic}, a prominent challenge in continual learning, which is usually addressed by utilizing shared representations to preserve previously learned information~\cite{pham2021dualnet, mallya2018packnet}.%

To analyze this, we take a three-modality alignment, \ie IMU, skeleton, and video from UTD-MHAD dataset~\cite{chen2015utd}, as an example. 
As depicted in Fig.\ref{fig:representation}, we utilize t-SNE to render the representation space of each modality visible. As evident in Figure \ref{subfig:before_align}, before alignment, features that have not undergone alignment training exhibit significant distribution differences. Fig.~\ref{subfig:joint_align} shows the conventional triplet alignment successfully bridge the modality gaps, aligning the three modalities. In contrast, the expandable network architecture within \name employs a sequence of two-modality alignment training phases as a replacement for the joint alignment. As illustrated in Fig.~\ref{subfig:our_align_1}, we initially align the IMU and skeleton modalities followed by the video modality, effectively bridging the modality gaps as well.

Our method is flexible regarding alignment order. Fig.~\ref{subfig:our_align_2} shows representations from each modality achieved by an alternately ordered network: first aligning skeleton and video, then IMU. Despite varying sequences, a common representation space is achievable. Further evaluation would be discussed in Section~\ref{subsubsec:growth_order}.

\section{Adaptive Training Strategy}
\label{sec:training_strategy}

We further propose our training strategies to optimally integrate the insights derived from the newly aligned modality during network growth. Specifically, we implement two strategies for the training of the concept alignment module and the prototype network, respectively.

For the training of the concept alignment module during network growth, we employ adaptive weighted contrastive training. The key of this design lies in dynamically adjust the proportion of proximity between modalities during the modal alignment process.

As per Equation~\ref{equ:adaptive}, the contrastive loss in aligning modality $\alpha$ and $\beta$ includes two parts: $L_{\alpha\leftarrow\beta}$, the loss when $\beta$ approximates $\alpha$, and $L_{\beta\leftarrow\alpha}$, the loss when $\alpha$ approximates $\beta$. We find reliable and unreliable modalities in various modality combinations and datasets. Naturally, modalities with robust encoders and abundant data are more reliable, so we expect less reliable ones to converge towards them. During network growth, careful updates are needed in the junction modality tower to add insights from the branch without disrupting aligned modalities. Hence, we integrate weights into Equation~\ref{equ:loss} as follows:
\vspace{-2mm}
\begin{equation}\label{equ:adaptive}
    L_{\alpha\beta}^M = \frac{w_{{\alpha\leftarrow\beta}}\cdot{L_{\alpha\leftarrow\beta}^M}+ w_{{\beta\leftarrow\alpha}}\cdot{L_{\beta\leftarrow\alpha}^M}}{2}, 
\end{equation}

where $M$ represents a batch randomly drawn from the dataset $E_{\alpha\beta}$, and $w_{{\alpha\leftarrow\beta}}$ and $w_{{\beta\leftarrow\alpha}}$ denote the normalized weights. Intuitively, we lean towards attributing a larger weight $w_{{\alpha\leftarrow\beta}}$ if modality $\alpha$ is deemed more reliable and established, while a smaller weight is assigned otherwise.

Identifying the appropriate weights presents a challenge. A static weighting scheme is suboptimal as each modality may differ in respect to data volume and quality, encoder proficiency, as well as the fresh insights and contributions it brings to the aligned modalities.
As such, we opt for a dynamic weighting strategy. Particularly, we employ gradients as an indicator to adaptively modify the weights,
\begin{equation}
    w_{\alpha\leftarrow\beta}^M = \frac{1}{\|\nabla_{\alpha\leftarrow\beta}^M({\Gamma_\alpha}, {\Gamma_\beta})\|},
\end{equation}

where $\nabla$ represents the accumulated gradients of all parameters within the concept alignment modules of the modality towers ${\Gamma_\alpha}$ and ${\Gamma_\beta}$ when computing the loss $L_{\alpha\leftarrow\beta}^M$ within the batch $M$. We calculate $w_{\beta\leftarrow\alpha}^M$ in a similar way. Then we normalize them as,
\begin{equation}
    w_{\alpha\leftarrow\beta}^M+w_{\beta\leftarrow\alpha}^M=1,
\end{equation}

\begin{figure}[t]
    \centering
    \includegraphics[width=0.85\linewidth]{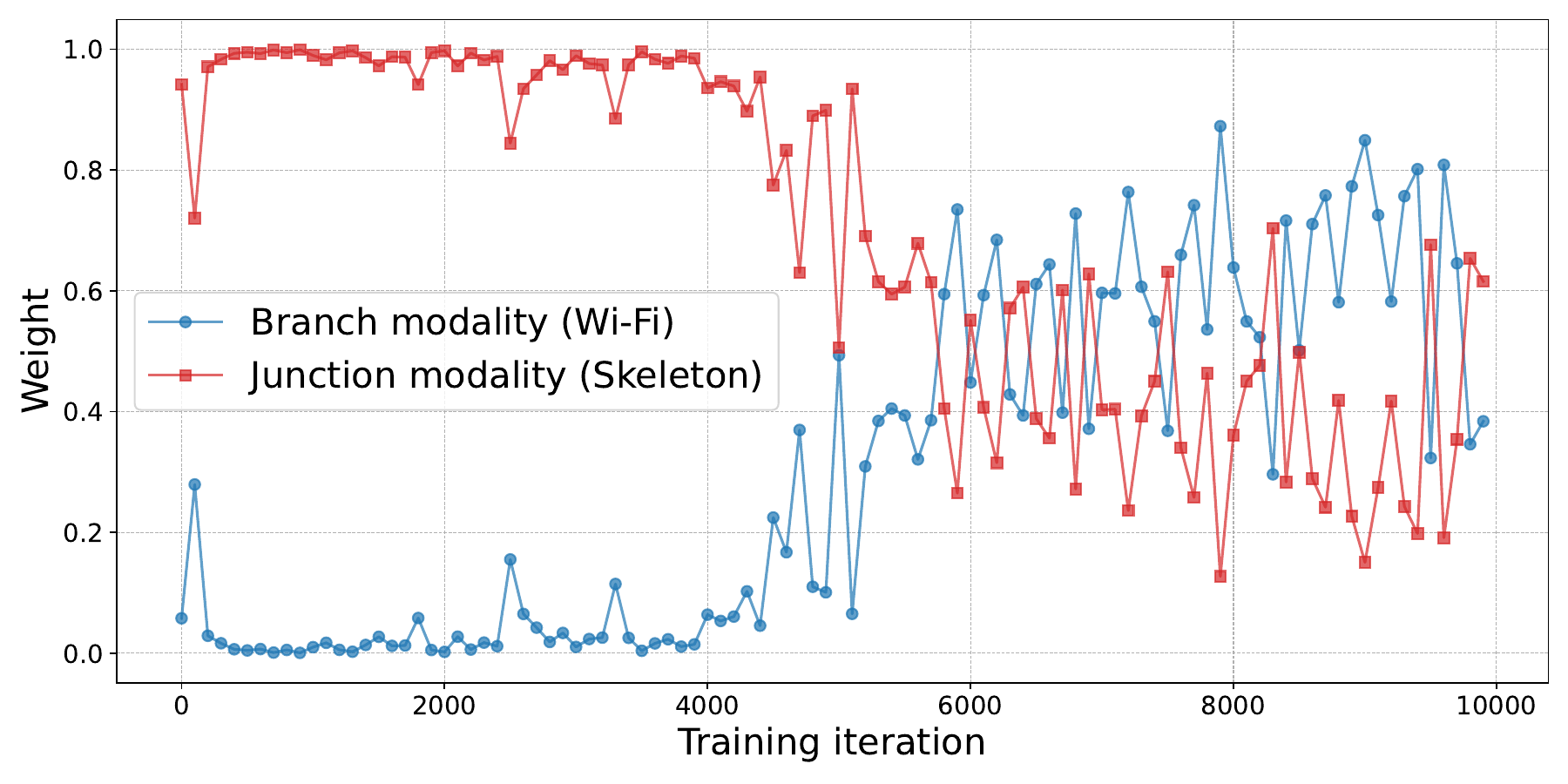}
    \caption{Adaptive training weights of branch and junction modality during the network growth.}
    \label{fig:adpative_weights}
\end{figure}

Gradient magnitudes effectively indicate how each modality contributes to the alignment process. During network growth, small gradients in the junction modality tower prompts a higher weight, bringing the branch network nearer to the trunk. When the branch modality tower's gradients are significant, the assigned weight speeds up the absorption of insights from the trunk network, ensuring alignment in the unified representation space. Our approach employs bidirectional contrastive learning for each pairwise alignment phase (e.g.,$\alpha\leftarrow\beta$ and $\beta\leftarrow\alpha$), but rather than maintaining fixed equal weights between these directions—which can lead to either insufficient adaptation of the branch modality or excessive perturbation of the trunk network—we dynamically adjust the update magnitudes for both modules by monitoring their gradient norms during training. When the branch exhibits larger gradients, indicating that the branch modality is farther from alignment, we increase the weight on the junction modality (pulling the new modality ``in''). Conversely, if the junction modality's gradient is large, this suggests that the new modality is relatively close, allowing the junction side to shift more than at the beginning. This adaptive weighting scheme naturally preserves previously learned representations and, through sufficient training iterations, ensures convergence to a consistent common representation, regardless of the order in which modalities are incorporated.

Fig.~\ref{fig:adpative_weights} shows the dynamic weight adaptation in the multi-modal alignment network construction using \name. This merges Wi-Fi as a branch modality into the trunk network, with the skeleton as the junction modality, using the OPERANet dataset for training~\cite{bocus2021operanet}. Initially, the skeleton modality, enriched with trunk network's aligned knowledge, is more reliable than the Wi-Fi branch modality, thus, it's assigned a near-one weight to speed up convergence with the junction modality. After around 6,000 training iterations, alignment is essentially achieved. Then, our dynamic weight adaptation mechanism adjusts to enable knowledge exchange between the junction and branch modalities, creating a comprehensive representation space.

For the training of the prototype network during network growth, we employ the exponential moving average (EMA) methodology. This strategy aids in preserving stability in the prototype representations by slowly incorporating fresh information while safeguarding the accumulated knowledge. We supplement this with knowledge distillation during the EMA process. This technique assists in preserving crucial information gleaned from prior modalities whilst incorporating novel ones.

\section{Implementations}
\label{sec:engineering}

\subsection{Data Preparation}

Overall, we utilize five datasets' training sets for the alignment, as itemized in Table~\ref{tab:datasets}, comprising paired samples across divergent dual modalities. These datasets are for human activity recognition (HAR) tasks, but the certain activities are totally different. Despite the provision of activity labels within these datasets, we adopt a self-supervised training approach, labels are not used. Throughout each dataset, we split into train and test set, detailed in Table~\ref{tab:datasets}.

\begin{table}[!t]
\centering
\caption{Datasets and their corresponding data pairs utilized to train the hexa-modal alignment network.}
    \begin{tabular}{l c r r}
    \hline
    \footnotesize{\textbf{Dataset}} & \footnotesize{\textbf{Modalities}} & \footnotesize{\textbf{\#  Train Pairs}}  & \footnotesize{\textbf{\# Test Pairs}} \\
    \hline
    \footnotesize{UTD-MHAD~\cite{chen2015utd}} & \footnotesize{IMU and Skeleton} & \footnotesize{$613$} & \footnotesize{$248$}\\
    \footnotesize{MM-Fi~\cite{yang2023mmfi}} & \footnotesize{LiDAR and Video}  & \footnotesize{$17,528$} & \footnotesize{$3,132$}  \\
    \footnotesize{OPERANet~\cite{bocus2021operanet}}  & \footnotesize{Wi-Fi and Skeleton}  & \footnotesize{$25,433$} & \footnotesize{$5,086$}  \\
    \footnotesize{XRF55~\cite{10.1145/3643543}}  & \footnotesize{mmWave and Wi-Fi}  & \footnotesize{$30,000$} & \footnotesize{$12,900$}  \\
    \footnotesize{Kinetics-400~\cite{kay2017kinetics}}  & \footnotesize{Video and Skeleton}  & \footnotesize{$234,619$} & \footnotesize{$19,761$}  \\
    \hline
    \end{tabular}
\label{tab:datasets}
\end{table}

\textbf{Skeleton\footnote{Depth signals undergo a conversion into a human skeleton format. As such, we employ the term \emph{skeleton} to denote the depth modality} and IMU pairs}. UTD-MHAD dataset~\cite{chen2015utd} is used, encompassing the skeleton and 9-axis IMU data pairs, captured via the Microsoft Kinect sensor and the wearable inertial sensor with respective sampling rates of 30Hz and 50Hz. The dataset embodies 27 distinct actions performed by 8 subjects. Each subject repeated the action for 4 times, totaling 861 paired samples. We use 613 pairs for the training.

\textbf{LiDAR and video pairs}. MM-Fi dataset~\citep{yang2023mmfi} is used, which contains 27 distinct actions performed by 40 human subjects. The LiDAR is collected in the point cloud format. MM-Fi~\citep{yang2023mmfi} dataset provides 17,528 pairs for our training.

\textbf{Wi-Fi and skeleton pairs}. OPERANet dataset~\citep{bocus2021operanet} is used, which contains the paired Wi-Fi CSI and skeleton data. The Wi-Fi CSI is gathered from the Intel 5300 platform across 30 subcarriers, employing a sampling rate of 1600Hz, with 3 transmitters and 3 receivers. The skeleton data is obtained from the Microsoft Kinect sensor. The dataset encompasses roughly 8 hours of annotated measurements collected in two different rooms with 6 participants performing 6 daily activities. OPERANet provides 25,433 pairs for our training.

\textbf{mmWave and Wi-Fi pairs}. XRF55 dataset~\citep{10.1145/3643543} is used, which is collected from a TI IWR6843ISK radar for mmWave and Intel 5300 for Wi-Fi CSI. It includes HAR data from 39 subjects performing 55 unique actions, each repeated 20 times. In total, 30,000 pairs are provided for training.

\textbf{Video and skeleton pairs}. Kinetics-400 dataset~\citep{kay2017kinetics} is used, which contains 400 distinct human action classes, each characterized by at least 400 video clips extracted from YouTube. Each clip, approximately 10 seconds long, portrays a variety of human actions. The skeleton is extracted from use clips using OpenPose~\citep{cao2017realtime}. Overall, as a dataset in vision modality, the Kinetics dataset provides 234,619 training pairs.

\subsection{Data Augmentation}

We implement two data augmentation techniques on the raw UTD-MHAD training data, ultimately enlarging the data pairs by 600$\times$. 
\textit{(i)} Down-sampling. Raw pairs undergo down-sampling at different ratios, simulating diverse sampling rates on various devices or accelerating the action at distinct ratios. This method augments the raw pairs by a factor of 300$\times$.
\textit{(ii)} Action-segmentation. The raw action sequence is randomly truncated, simulating incomplete activity sensing. We ensure the segmented sequence's shortest length is over 50\% of the original length. This method amplifies the raw pairs by a factor of 300$\times$.

\subsection{Pre-trained Encoders and Concept Alignment Architecture}

Next we introduce the pre-trained encoders we use for building the modality alignment network, along with the concept alignment architecture.

\textbf{IMU}. We utilize the LIMU-BERT encoder~\citep{xu2022limu}, renowned for its proficiency in generating generalized representations. 
It is pre-trained on a range of IMU datasets.

\textbf{Skeleton}. We utilize the Spatial-Temporal Graph Convolutional Network (ST-GCN)~\citep{yan2018spatial} as our encoder, which is pre-trained on extensive datasets, notably the NTU-RGBD~\citep{shahroudy2016ntu}.

\textbf{Video}. We employ ResNet3D model ~\citep{tran2018closer} as the encoder, which is pre-trained on Kinetics-400 dataset~\cite{kay2017kinetics}.

\textbf{Wi-Fi}. For Wi-Fi CSI, we fails to obtain one powerful pre-trained encoder. Therefore we apply multiple encoders to augment the modality tower of Wi-Fi. Specifically, we utilized a Vision Transformer (ViT) and a combination of Convolutional Neural Network (CNN) and Gated Recurrent Unit (GRU) as our encoders. They are pre-trained on UT-HAR~\citep{yousefi2017survey} datasets.

\textbf{mmWave}. We employ the signal processing based encoder for this modality. We use doppler fast fourier transform (FFT) and angle FFT, generating range-doppler heatmaps and range-angle heatmaps, respectively. We supply an additional spatial ResNet18~\citep{10.1145/3643543} to further extract features from them. %

\textbf{LiDAR}. We use the Point Transformer~\citep{DBLP:journals/corr/abs-2012-09164}, which is pre-trained on the ModelNet40 dataset~\citep{wu20153d}. The encoder cannot extract temporal features, we add an additional ST-GCN as the additional temporal feature extractor, which is pre-trained on the NTU-RGBD~\citep{shahroudy2016ntu} dataset.

The \textbf{concept alignment architecture} comprises two Multi-Layer Perceptrons (MLPs) for each modality. This concept alignment module is customized for each modality, with the input dimensionalities determined by their encoders: LiDAR (60), Skeleton (60), IMU (72), Video (512), Wi-Fi (900), and mmWave (1024).

\subsection{Training Details}

We commence the training process with the IMU and skeleton modalities. Subsequently, we integrate the video modality, aligning it with the pre-existing skeleton modality. Next, we incorporate the Wi-Fi modality into our framework, leveraging the paired Wi-Fi and skeleton data. This is followed by the introduction of the mmWave modality, which is linked with the intermediate Wi-Fi modality. Ultimately, we incorporate the LiDAR modality, capitalizing on its integration with the paired video modality.

We employ the AdamW optimizer~\citep{loshchilov2019decoupled} with a batch size of 256 and an initial learning rate of $1\times10^{-4}$. Given a batch size of $m = 256$, we construct 256 positive pairs and $65,280$ negative pairs for contrastive learning, following the construction method detailed in Section~\ref{subsec:align}. For each phase of network growth, we judiciously allocate a varying number of training epochs, typically up to 500, or cease the training process once convergence is attained. The learning rate for downstream tasks is adjusted between 0.001 and 0.1. We train on two NVIDIA A100 GPUs, spending around 20 hours to align six modalities.

\section{Evaluation}
\label{sec:evaluation}

We evaluate pre-trained \name by employing a typical downstream sensing task, human activity recognition (HAR). Furthermore, we would demonstrate two applications enabled by \name: cross-modality retrieval and LLM integration.

\subsection{Evaluation on HAR}

We evaluate \name on 8 datasets, comprising 4 in-domain datasets including UTD-MHAD~\cite{chen2015utd}, OPERANet~\cite{bocus2021operanet}, XRF55~\cite{10.1145/3643543} and MM-Fi~\cite{yang2023mmfi}. We utilize the test pairs outlined in Table~\ref{tab:datasets}. Additionally, we assess on 4 out-of-domain datasets, which were not part of the pre-training datasets at all: UCI~\citep{reyes2016transition}, Widar3.0~\citep{9516988}, mRI~\cite{an2022mrimultimodal3dhuman} and MSRAction3D~\cite{li2010action}. Out-of-domain evaluation is critical for understanding model generalization in real-world sensing applications. In practical deployments, sensing models frequently encounter new environments and previously unseen subjects or activities. By evaluating on out-of-domain datasets, we can measure how well \name adapts to new scenarios without any additional training.

The results for \name are obtained in a \emph{one-shot} setting, which serves as a widely used benchmark for evaluating the generalization capability of a backbone model~\cite{fei2006one, koch2015siamese}. Unlike the conventional fully supervised scenario, one-shot learning is particularly relevant for sensing applications, where labeled samples are expensive and difficult to acquire~\cite{xiao2021onefi, feng2019few}. In this setup, we select only \textit{one labeled} sample per class from the test portion for fine-tuning on the downstream task, while the remaining test samples are used to compute the accuracy metrics reported in our results tables. This challenging setting underscores \name's effectiveness as a pre-trained network, demonstrating its ability to generalize with minimal supervision in resource-constrained sensing environments.

We compare \name with a broad range of baselines, including SOTA singular-modal sensing baselines, LIMU-BERT~\citep{xu2021limu} for IMU, SenseFi~\citep{yang2023sensefilibrarybenchmarkdeeplearningempowered} for Wi-Fi, MARS~\citep{an2021mars} for mmWave, MeteorNet~\citep{liu2019meteornet} together with PointTransformer~\citep{DBLP:journals/corr/abs-2012-09164} for LiDAR. Additionally, we include the multi-modal sensing baseline Cosmo~\cite{ouyang2022cosmo}. Finally, we compare \name with emerging MLLMs that hold potential for interpreting sensing signals, including OneLLM~\cite{han2023onellm} and M4~\citep{M4mengwei}.

\begin{figure}[t]
    \centering
    
    \begin{subfigure}{0.48\linewidth}
     \includegraphics[width=\linewidth]{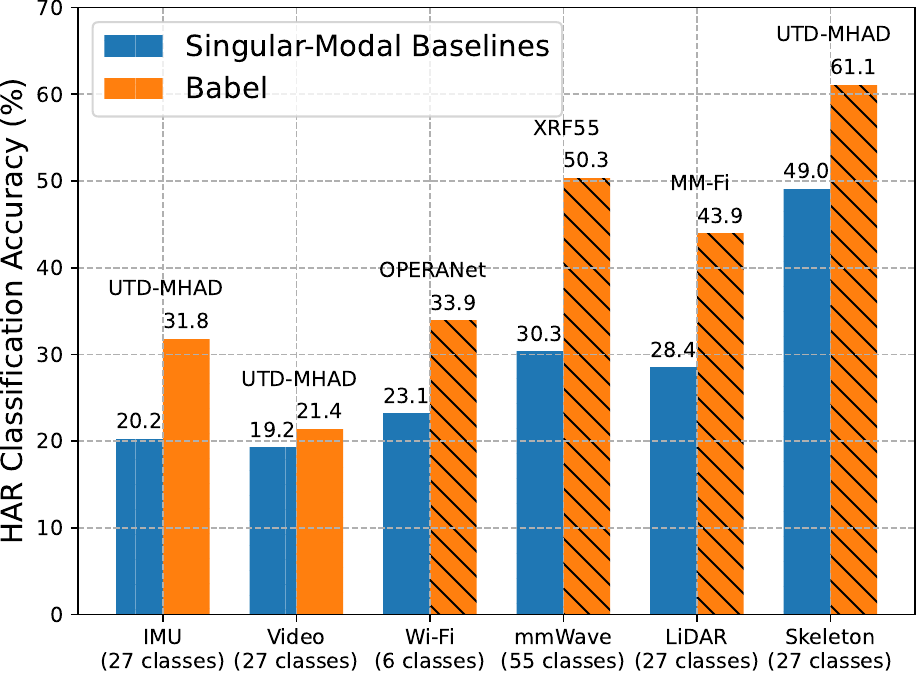}
     \caption{On in-domain datasets}
     \label{fig:eva_single_in_domain}
    \end{subfigure}
    \hfill
   \begin{subfigure}{0.48\linewidth}
    \includegraphics[width=\linewidth]{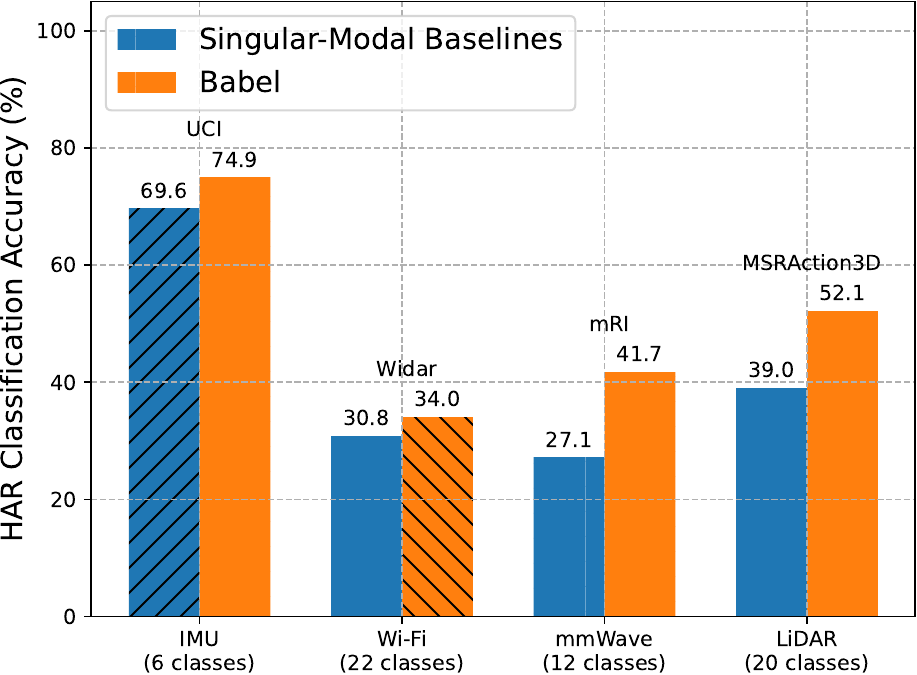}
    \caption{On out-of-domain datasets}
    \label{fig:eva_single_out_domain}
    \end{subfigure}
    \vspace{-1em}
    \caption{One-shot HAR classification accuracy of each modality achieved by \name on the in-domain and out-of-domain datasets compared to various singular-modal baselines: LIMU-BERT(IMU), SenseFi(Wi-Fi)~\cite{yang2023sensefi}, MARS(mmWave)~\cite{an2021mars}, PointTransformer(LiDAR), ResNet3D(Video), ST-GCN(Skeleton). For out-of-domain datasets, \name achieves 4.5×, 7.6×, 5×, and 10.4× improvements over random guessing (IMU: 16.7\%, Wi-Fi: 4.5\%, mmWave: 8.3\%, LiDAR: 5.0\%).
}
\end{figure}

\subsubsection{Performance on singular-modal sensing.}

Owing to the pre-trained alignment across multiple sensing modalities, \name exhibits superior performance even when each individual aligned modality is applied to downstream tasks. In our evaluation, we apply one‐shot training and testing exclusively to each modality’s data—even if the dataset itself contains multiple modalities. For instance, although UTD‐MHAD includes IMU, video, and skeleton data for the same set of actions, we use only the IMU samples for one‐shot training and testing on the IMU, only the video samples for the video evaluation, and so forth. Fig.~\ref{fig:eva_single_in_domain} presents the evaluation results of singular-modal sensing across four in-domain datasets. As illustrated, compared to SOTA singular-modal methods, \name delivers an average accuracy improvement of approximately 12\% across six aligned modalities in various datasets. Notably \name achieves significant gains in weaker sensing modalities. For instance, classification accuracy for 27 human activities in the IMU modality increases from 20.19\% to 31.77\%, compared to LIMU-BERT on UTD-MHAD. 
The Wi-Fi modality obtains an approximate 10.74\% enhancement. The mmWave modality shows a substantial increase from 30.32\% to 50.30\%, and the LiDAR modality achieves an accuracy of 43.91\%, up from 28.43\%. 
Such gains are achieved by aligning each modality into a unified representation space, facilitating mutual learning from strong modalities, \ie video. Though sensing modalities benefit significantly, gains are limited for the video modality, increasing by around 2\%.

The performance of \name on four full out-of-domain datasets is detailed in Fig.~\ref{fig:eva_single_out_domain}. Owing to the effectiveness of multi-modality alignment, \name consistently outperforms the SOTA methods for each individual modality. Notably, \name demonstrates significant improvements for the mmWave and LiDAR modalities, achieving gains of 14.6\% and 13.1\%, respectively. In the Wi-Fi modality, \name outperforms SenseFi by 5.3\%. For the IMU modality, \name attains an accuracy of 74.9\%.
It is important to note that none of the datasets evaluated here were included in the pre-training dataset, highlighting the generality of \name.

\begin{table}[!t]
\centering
\caption{The classification accuracy of sensing fusion for \name and the multi-modal sensing baselines evaluated on the out-of-domain mRI\citep{an2022mrimultimodal3dhuman} dataset.} 
\begin{adjustbox}{width=\linewidth}
\begin{tabular}{c c c}
\hline

{} & \textbf{Multi-modal Baselines} & \textbf{Babel}  \\
\hline
\textbf{Vision+IMU} & {$89.6\%$} & {$91.7\%$(+$2.3\%$)}\\
\textbf{Vision+mmWave} & {$58.4\%$} & {$64.6\%$(+$10.7\%$)}\\
\textbf{IMU+mmWave} & {$75.0\%$} & {$86.5\%$(+$13.5\%$)}\\
\textbf{Vision+IMU+mmWave} & {$85.4\%$} & {$92.8\%$(+$8.6\%$)}\\
\hline
\end{tabular}
\end{adjustbox}
\label{tab:eva_out_multi_modal}
\end{table}

\subsubsection{Performance on multi-modal sensing.} The unified representation space in \name allows for effective fusion. When IMU and Video modalities are fused, \name achieves a 33.17\% accuracy on UTD-MHAD~\cite{chen2015utd}, outperforming both the individual IMU and video modalities. Likewise, a 58.97\% accuracy is achieved on XRF55~\cite{10.1145/3643543} when Wi-Fi and mmWave modalities are merged. Note that even modality combinations (like IMU\&Video fusion) not included in pre-training datasets are evaluated and obtain the superior performance, highlighting \name's flexibility and offering developers numerous opportunities to choose any one or combined modalities for their tasks.

We also evaluate the multi-modal sensing fusion of \name on an out-of-domain dataset, mRI\citep{an2022mrimultimodal3dhuman}, which is not included in the pre-training datasets. Various combinations of modalities are evaluated, and \name's performance is compared with multi-modal sensing baselines. These baselines are implemented following the designs of existing works~\citep{an2021mars, an2022mri}. As shown in Table~\ref{tab:eva_out_multi_modal}, \name achieves up to a 13.5\% improvement in accuracy compared to the modality-specific baselines.

\begin{table}[!t]
\centering
\caption{Comparison with Cosmo \citep{ouyang2022cosmo} under various configurations for IMU-Skeleton fusion sensing: \name(B) represents the \name network that only aligns two modalities, while Cosmo(G) denotes the network customized its specific structure with simple MLPs.} 
\begin{adjustbox}{width=\linewidth}
\begin{tabular}{c c c c c}
\hline
{} & \textbf{Babel(B)} & \textbf{Cosmo} & \textbf{Cosmo(G)} & \textbf{Babel} \\
\hline
\textbf{Aligned Modalities} & 2 & 2 & 2 & 6 \\
\hline
\textbf{Downstream} & \multirow{2}{*}{MLP} & Specific network & \multirow{2}{*}{MLP} & \multirow{2}{*}{MLP}\\
\textbf{Task Designs} & & and training strategy & \\
\hline
\textbf{IMU-Skeleton} & \multirow{2}{*}{61.46\%} & \multirow{2}{*}{56.3\%} & \multirow{2}{*}{41\%} & \multirow{2}{*}{63.02\%} \\
\textbf{Fusion Acc.}\\
\hline
\end{tabular}
\end{adjustbox}
\label{tab:compare_with_cosmo}
\end{table}

\subsubsection{Comparison with Cosmo~\citep{ouyang2022cosmo}} Cosmo is the SOTA sensing fusion framework, but unlike \name, it requires all modalities to coexist within one dataset, limiting its expandability to datasets with complete paired data.
Thus to compared with Cosmo, we utilize the same paired IMU-skeleton from the UTD-MHAD~\cite{chen2015utd} that it excels.
An equal amount of data is employed to train both Cosmo and a bi-modality version of \name. 
We train Cosmo and \name for 10 times with different random seeds. In the fusion of IMU and skeleton modalities, Cosmo achieves an averaged classification accuracy of 56.3\%, while \name attains 61.46\% on UTD-MHAD, as shown in Table~\ref{tab:compare_with_cosmo}.

What's more, Cosmo's performance can also be attributed to the integration of an additional network structure and its corresponding training procedure (for each downstream task), referred to as iterative fusion learning. Conversely, we aim to highlight \name's efficacy as a pre-trained network with a simple downstream task design. When applying the same downstream network (\ie MLP), \name could achieve around 20\% accuracy improvement compared to Cosmo. Furthermore, as a expandable solution, \name allows aligning more modalities without retraining pre-existing ones. This enhances \name's performance when introducing modalities. As shown, \name aligning six modalities during the pre-training phase could achieve 63.02\% on the IMU-Skeleton fusion sensing task.

\begin{table}[!t]
\centering
\caption{HAR classification accuracy with \name and typical MLLMs on UTD-MHAD datasets.} 
\begin{adjustbox}{width=\linewidth}
\begin{tabular}{c c c c c c c}
\hline
\multirow{2}{*}{\textbf{MLLMs}} &  \textbf{IMU} & \textbf{Video} & \textbf{Wi-Fi} & \textbf{mmWave} & \textbf{Skeleton} & \textbf{LiDAR}\\
& UTD-MHAD~\cite{chen2015utd} & UTD-MHAD & OPERANet~\cite{bocus2021operanet} & XRF55~\cite{10.1145/3643543} & UTD-MHAD & MM-Fi~\cite{yang2023mmfi}\\
\hline
\textbf{OneLLM}~\cite{han2023onellm} with & \multirow{2}{*}{$6.5$\%} & \multirow{2}{*}{$6.51$\%} & \multirow{2}{*}{$-$}  & \multirow{2}{*}{$-$} & \multirow{2}{*}{$-$} & \multirow{2}{*}{$-$}\\
\textbf{Meta-Transformer}~\citep{zhang2023metatransformer} \\
\hline
\textbf{M4}~\cite{M4mengwei} with & \multirow{2}{*}{$5.77$\%} & \multirow{2}{*}{$7.44$\%} & \multirow{2}{*}{$-$}  & \multirow{2}{*}{$-$} & \multirow{2}{*}{$-$} & \multirow{2}{*}{$-$}\\
\textbf{ImageBind}~\citep{girdhar2023imagebind}\\
\hline
\textbf{Ours} & $\textbf{31.77}$\% & $\textbf{21.35}$\% & $\textbf{33.89}$\%  & $\textbf{50.30}$\% & $\textbf{61.06}$\% & $\textbf{43.91}$\%\\
\hline
\end{tabular}
\end{adjustbox}
\label{tab:performance_mllms}
\end{table}

\subsubsection{Comparison with MLLMs.} There has been a significant development in MLLMs~\citep{han2023onellm, zhang2023metatransformer}. These models are capable of understanding multi-modal inputs, including sensing modalities like IMU potentially. For comparison, we select typical MLLMs \eg OneLLM~\cite{han2023onellm}and M4~\cite{M4mengwei}, and evaluate their performance on the UTD-MHAD~\cite{chen2015utd} with HAR tasks. OneLLM uses Meta-Transformer~\cite{zhang2023metatransformer} and M4 uses ImageBind~\cite{girdhar2023imagebind} to interpret sensing signals, respectively. The results are summarized in Table~\ref{tab:performance_mllms}. Firstly, current MLLMs can only support a limited number of sensing modalities, \ie IMU. Secondly, they only achieve a classification accuracy of around 5\%-6\% according to our evaluation. In stark contrast, \name significantly outperforms these with a classification accuracy of 31.77\% on IMU while supporting other five sensing modalities.

The rationale that these MLLMs seems supporting sensing modalities, but struggle to comprehend IMU data and manage HAR tasks, is their training only on the Ego4D dataset~\citep{grauman2022ego4d}. Without sufficient training, these models are restricted to trained data, limiting their cross-domain capabilities. Furthermore, these MLLMs are unable to be trained on other sensing datasets due to data scarcity and absence of techniques like pre-trained modality tower and the expandable architecture, which are introduced in \name. MLLMs could be improved by incooperating \name as the sensing modality encoders, which would be discussed in \S\ref{subsec:case_study}

\subsubsection{Growth Orders}\label{subsubsec:growth_order} Thanks to the prototype network architecture and the adaptive training strategy proposed in \name, the order of modality growth does not significantly influence the end-to-end performance  once the training is sufficient. To evaluate this, we devise four network growth sequences according to different heuristics: \textit{(i)} random order: the modalities are aligned in the sequence of IMU, skeleton, video, Wi-Fi, mmWave, and LiDAR. \textit{(ii)} alignment from the most robust to weakest modality (skeleton, video, LiDAR, IMU, Wi-Fi, mmWave); \textit{(iii)} alignment based on data diversity, taking into account the number of actions, subjects, and scenes of the used datasets; The sequence follows skeleton, mmWave, LiDAR, IMU, and Wi-Fi. \textit{(iv)} alignment based on the data amount of used datasets, organized from largest to smallest, proceeding from skeleton, mmWave, Wi-Fi, LiDAR, and IMU. Using these different growth orders, we train \name and evaluate the end-to-end classification accuracy on the downstream tasks. As shown in Table~\ref{tab:performance_align_orders}, growth order doesn't significantly affect the performance. For instance, with different growth orders, the performance on IMU and Wi-Fi modality varies less than 3\% and 2\%, respectively. This highlights \name's robustness.

\subsubsection{Ablation} The techniques proposed in \name, including the pre-trained encoders, expandable network architecture and adaptive training strategy, are all essential for constructing \name. Particularly, Without pre-trained modality tower, training wouldn't converge due to limited samples. On UTD-MHAD~\cite{chen2015utd}, without prototype network, the previously aligned modality would drop about averagely 44.7\% relatively after introducing a new modality. Without adaptive training, the overall performance would decrease by up to 7.2\%.

\begin{table}[!t]
\centering
\caption{Performance of each modality when applying different heuristics to determine growth orders.}
\begin{adjustbox}{width=\linewidth}
\begin{tabular}{c c c c c c c}
\hline
\multirow{2}{*}{\textbf{Heuristics}} & \textbf{IMU} & \textbf{Skeleton} & \textbf{Video} & \textbf{Wi-Fi} & \textbf{mmWave} & \textbf{LiDAR}\\
& UTD-MHAD~\cite{chen2015utd} & UTD-MHAD & UTD-MHAD & OPERANet~\cite{bocus2021operanet} & XRF55~\cite{10.1145/3643543} & MM-Fi~\cite{yang2023mmfi}\\
\hline
\textbf{Random}         & $31.77$\% & $61.06$\% & $21.35$\% & $33.89$\% & $50.30$\% & $43.91$\% \\ 
\textbf{Robustness} & $29.33$\% & $60.58$\% & $20.83$\% & $35.31$\% & $52.16$\% & $44.65$\% \\ 
\textbf{Diversity} & $27.60$\% & $56.25$\% & $21.35$\% & $35.79$\% & $47.93$\% & $44.21$\% \\ 
\textbf{Amount}    & $28.13$\% & $59.90$\% & $20.83$\% & $33.85$\% & $46.85$\% & $47.70$\% \\ 
\hline
\end{tabular}
\end{adjustbox}
\label{tab:performance_align_orders}
\end{table}

\begin{figure}[t]
    \centering
    \includegraphics[width=0.9\linewidth]{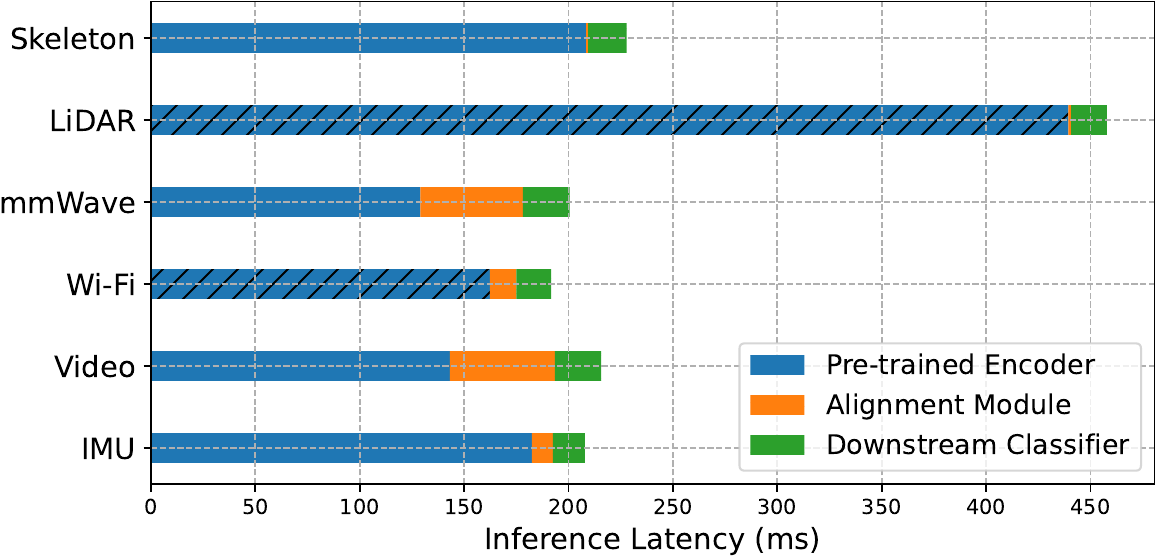}
    \caption{Per-sample inference latency breakdown for each modality in \name evaluated on NVidia A100 GPU.}
    \label{fig:eva_system_overhead}
\end{figure}

\subsubsection{System Overhead.} In \name, we demonstrates efficiency by introducing minimal additional system overhead. Fig.~\ref{fig:eva_system_overhead} illustrates the breakdown of inference latency for each modality in \name, evaluated on a NVidia A100 GPU for an activity sample. As depicted, the overhead introduced by \name, specifically the alignment module, is very limited compared to the pre-trained encoders. For example, in the IMU modality, the alignment module requires approximately 10.1 ms, while the pre-trained encoder, LIMU-BERT, takes 182.4 ms to process a fixed window of IMU signals. On average, the alignment module incurs only about an 8\% increase in inference overhead based on our evaluation. The overhead of the alignment module varies across modalities due to the use of different MLP layer configurations to accord with each modality's encoder. Aside from the alignment module, the prototype network is shared among modalities, with its overhead being negligible, less than 1 ms. When applying multiple modalities to downstream tasks, each modality tower can be parallelized to hide multi-modal sensing latency.

The pre-trained weights of \name occupy approximately 1.1GB on disk, encompassing pre-trained encoders for all modalities, concept alignment modules, and the prototype network. Depending on the selected modalities, \name requires 1.4-9.92GB of memory using FP32 precision.

\begin{figure}[!t]
\centering
\includegraphics[width=0.95\linewidth]{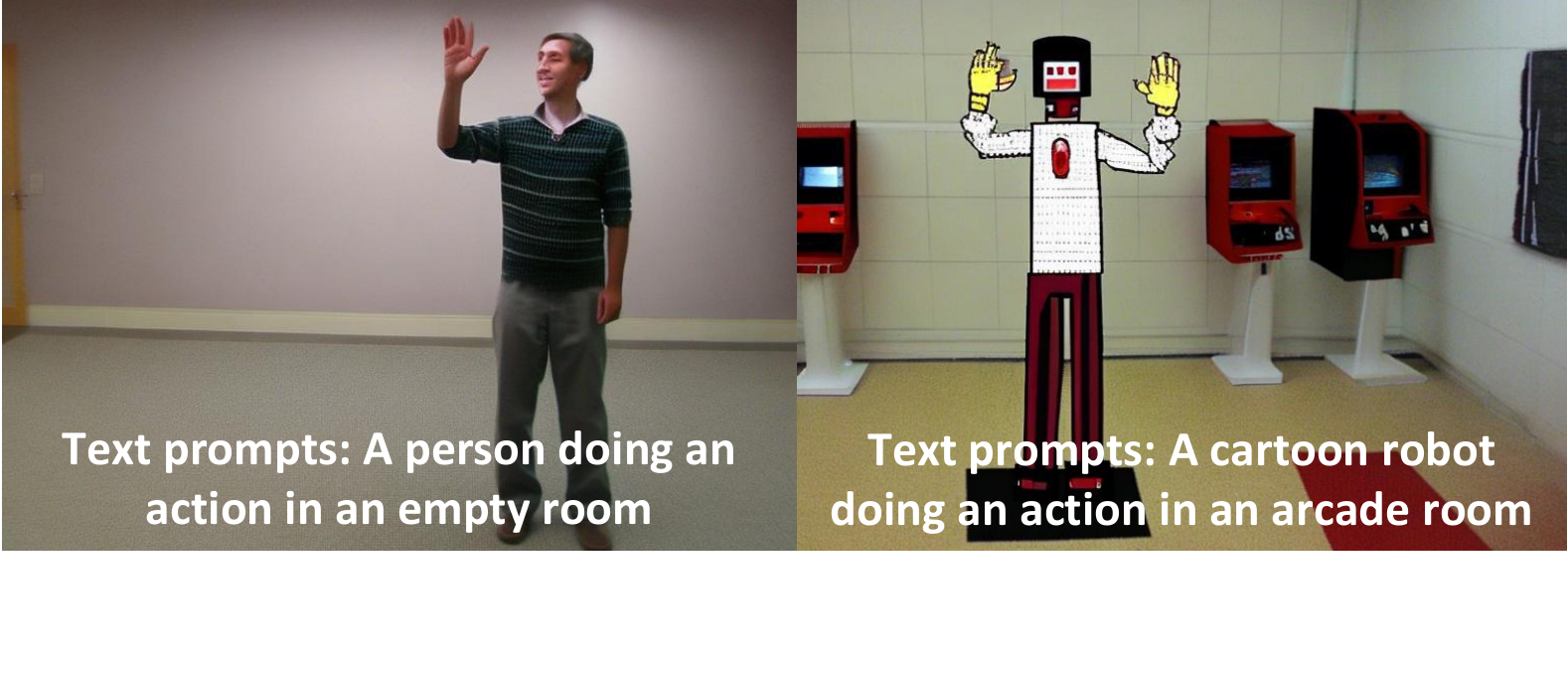}
\vspace{-2em}
\caption{Images generated through cross-modality retrieval. The action information (waving hands) is input via IMU, the environment information is input through text prompts.}
\vspace{-1em}
\label{fig:test}
\end{figure}

\subsection{Case Study}
\label{subsec:case_study}
\subsubsection{Cross-modality retrieval}

The alignment of diverse sensing modalities in \name potentially opens up the possibilities for cross-modality retrieval applications. This involves obtaining the representations of one modality using signals from other modalities as inputs. Such applications could be promising. For instance, using wireless sensing signals as input to retrieve visual representations could be considered an example of \emph{sensing imaging}.

To showcase, we construct a prototype designed to retrieve visual representations and generate images using non-visual sensors, such as IMU. Specifically, we align \name with unCLIP~\citep{ramesh2022hierarchical}, an image-to-image diffusion model. unCLIP employs an image encoder to obtain the embeddings of the input image and then uses these embeddings to guide the diffusion process, thereby generating images that bear stylistic similarities to the input image. We incorporate unCLIP's image encoder into our \name network, enabling the sensing modalities to be interpreted by the diffusion module in unCLIP. We use L1 loss to align \name and unCLIP.

Fig.~\ref{fig:test} demonstrates the images generated using IMU as inputs, representing the sensor readings of a person gesturing with hands. Leveraging unCLP, the actions captured by IMU are visually represented. The environmental information and other visual styles are provided through text prompts.
We believe this area of research opens up interesting possibilities, offering a pathway to visualizing the physical world through non-visual sensors.

\subsubsection{Bridge with LLMs.}

The alignment of diverse sensing modalities into a unified representation presents an advantageous prospect for integration with LLMs. 
To demonstrate, we integrate \name with Video-LLaMA~\citep{zhang2023videollama}, which is a multi-modal LLM with the ability to understand both visual and audio contents.

We establish the alignment between the video modality in \name and that in Video-LLaMA. Specifically, we judiciously select the video encoder from Video-LLaMA and construct a modality tower for integration into \name. We employ the L1 loss in this scenario, ensuring the video encoder of Video-LLaMA remains frozen while all modalities in \name align towards Video-LLaMA. This strategy aims to generate embeddings of sensing modalities that could potentially be interpreted by Video-LLaMA.

\begin{figure}[!t]
    \centering
    \includegraphics[width=0.9\linewidth]{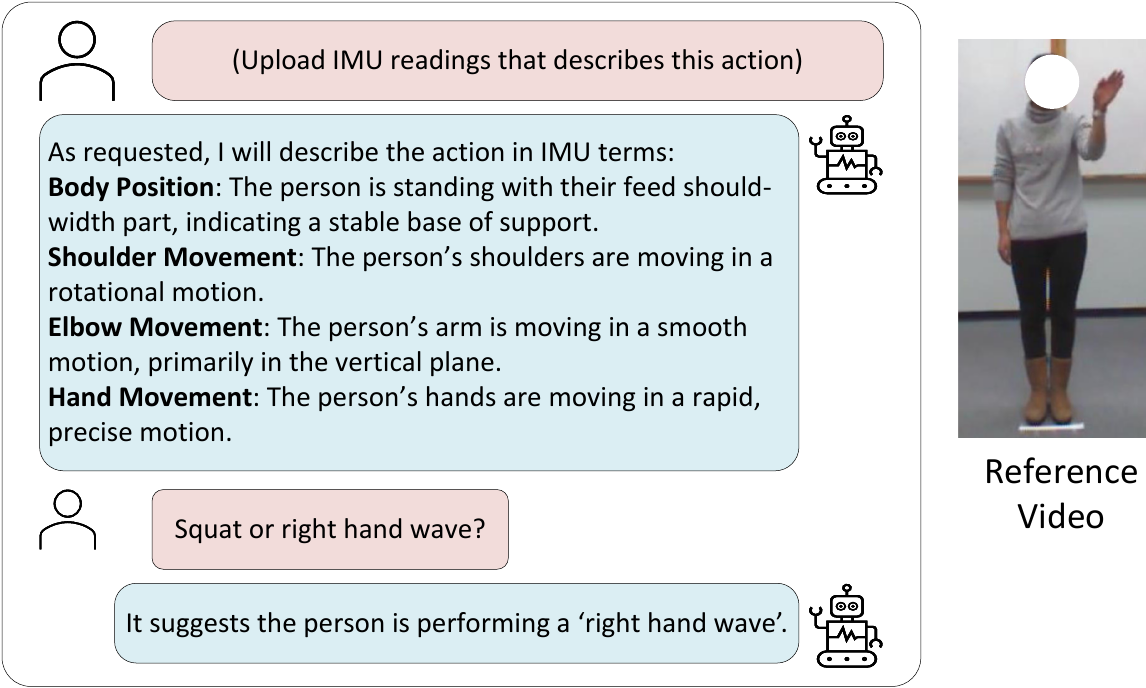}
    \caption{With \name, Video-LLaMA accepts IMU readings as inputs and conduct a preliminary analysis of the actions represented by these IMU readings.}
    \label{fig:llm}
\end{figure}

Fig.~\ref{fig:llm} provides an impressive illustration where we input an IMU sequence depicting a woman waving her hands. These IMU readings are processed by \name and subsequently fed into Video-LLaMA. Remarkably, without any specific training on LLMs, it successfully deciphers the action captured by the IMU data and, when promoted, differentiates between diverse actions, such as squatting or waving hands. This exemplifies the potential of bridging sensing and LLMs via the modality alignment introduced by \name. 
Our future research will concentrate on improving \name, aiming to bolster the model's capability to provide deeper insights and more accurate interpretations of physical world based on a broader spectrum of sensing modalities, and bringing such the capabilities to LLMs.

\section{Discussion and Future Work}

While \name demonstrates promising results in aligning multiple sensing modalities for HAR tasks, several directions remain for our future exploration. Firstly, the framework could be extended to other sensing applications including localization, gesture detection~\cite{sun2020real}, and autonomous navigation. Secondly, Our preliminary exploration of integrating \name with LLMs demonstrates the potential for enhanced sensing comprehension. The current approach of aligning sensing embeddings with video encoders in MLLMs. Future research should investigate direct integration methods that can map sensing features into LLM's native representation spaces without requiring intermediate alignment steps. Our case studies in sensing imaging and LLM integration could be further benefit from more state-of-the-art diffusion models~\cite{wu2023tuneavideooneshottuningimage, musev, hu2024animateanyoneconsistentcontrollable} and LLMs~\cite{lin2023video,zhang2024internlm, li2024llavaonevisioneasyvisualtask}.

\section{Conclusion} %

We present \name, a expandable modality alignment model designed for sensing applications. The pre-trained \name has been proficiently aligned with six prevalent sensing modalities, IMU, skeleton, video, Wi-Fi, LiDAR, and mmWave. \name demonstrated the superior performance for HAR tasks across various datasets compared to an array of baselines. As \name is a scalable network, we call for the community to further enhance and align additional helpful modalities into \name.

\balance
\bibliographystyle{ACM-Reference-Format}
\bibliography{ref}

\end{document}